\documentclass{opendatalab}

\usepackage[utf8]{inputenc}
\usepackage{xcolor}
\usepackage[most]{tcolorbox}
\usepackage{longtable}
\usepackage{listings}
\usepackage[numbers]{natbib}
\usepackage{enumitem}
\usepackage{graphicx}
\usepackage{xspace} 
\usepackage{hyperref}
\usepackage{xspace}

\makeatletter
\DeclareRobustCommand\onedot{\futurelet\@let@token\@onedot}
\def\@onedot{\ifx\@let@token.\else.\null\fi\xspace}

\def\eg{\emph{e.g}\onedot}

\makeatother
\definecolor{codegreen}{rgb}{0,0.6,0}
\definecolor{codegray}{rgb}{0.5,0.5,0.5}
\definecolor{codepurple}{rgb}{0.58,0,0.82}
\definecolor{backcolour}{rgb}{0.95,0.95,0.92}
\definecolor{promptcolor}{HTML}{D1D0F2}
\definecolor{promptcolorheader}{HTML}{bdbcec}

\newcommand{\github}{\raisebox{-1.5pt}{\includegraphics[height=1.05em]{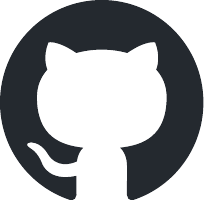}}\xspace}
\newcommand{\web}{\raisebox{-1.5pt}{\includegraphics[height=1.05em]{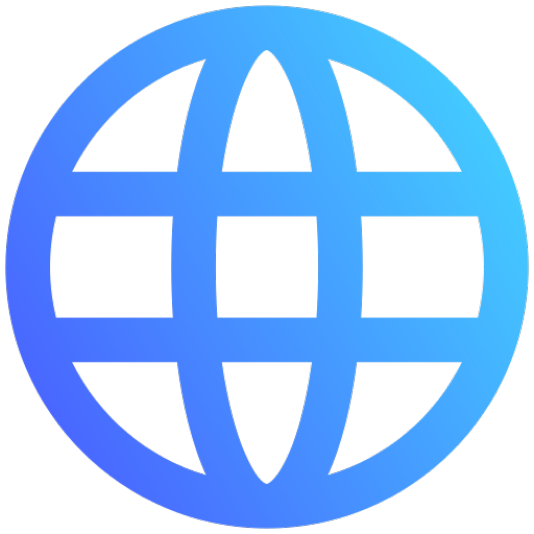}}\xspace}
\newcommand{\huggingface}{\raisebox{-1.5pt}{\includegraphics[height=1.05em]{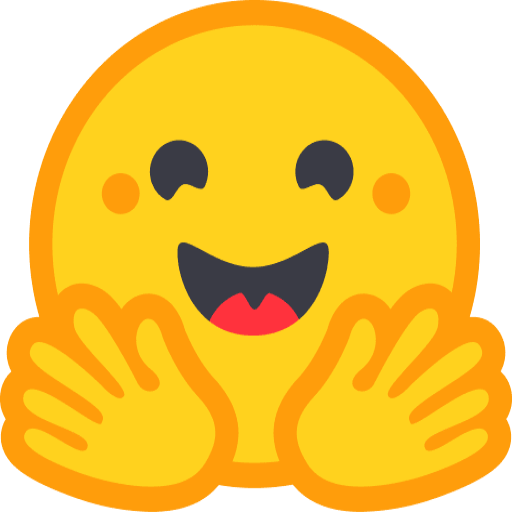}}\xspace}

\definecolor{promptcolor}{HTML}{D1D0F2}
\definecolor{promptcolorheader}{HTML}{bdbcec}

\newtcolorbox{promptbox}[1][]{
  enhanced, breakable,
  top=0.3em,bottom=0.3em,left=0.5em,right=0.5em,
  toptitle=0.3em,bottomtitle=0.2em,boxsep=0pt,
  colframe=promptcolorheader, colback=promptcolor!50, boxrule=0.5pt,
  width=\columnwidth, 
  title={\footnotesize #1} 
}
\lstdefinestyle{promptstyle}{
    backgroundcolor=\color{backcolour},   
    commentstyle=\color{codegreen},
    keywordstyle=\color{magenta},
    numberstyle=\tiny\color{codegray},
    stringstyle=\color{codepurple},
    basicstyle=\ttfamily\footnotesize,
    breakatwhitespace=false,         
    breaklines=true,                 
    captionpos=b,                    
    keepspaces=true,                 
    numbers=left,                    
    numbersep=5pt,                  
    showspaces=false,                
    showstringspaces=false,
    showtabs=false,                  
    tabsize=2
}
\title{PointCoT: A Multi-modal Benchmark for Explicit 3D Geometric Reasoning}

\author{
    \parbox{\linewidth}{\raggedright
        Dongxu Zhang$^{1,9,*}$ \quad
        Yiding Sun$^{1,9,*}$ \quad
        Pengcheng Li$^{2,9,*}$ \quad
        Yumou Liu$^{3}$ \quad
        Hongqiang Lin$^{4}$ \\ \vspace{0.15cm} % 增加了微调行距，防止上下角标打架
        Haoran Xu$^{1}$ \quad
        Xiaoxuan Mu$^{1}$ \quad
        Liang Lin$^{5}$ \quad
        Wenbiao Yan$^{6}$ \quad
        Ning Yang$^{7}$ \\ \vspace{0.15cm}
        Chaowei Fang$^{1}$ \quad
        Juanjuan Zhao$^{8}$ \quad
        Jihua Zhu$^{1}$ \quad
        Conghui He$^{9}$ \quad
        Cheng Tan$^{9}$
    }
}

\affiliation{
    \vspace{0.2cm} % 与作者列表保持呼吸感
    \parbox{\linewidth}{\raggedright \small
        $^1$Xi'an Jiaotong University \quad
        $^2$Tsinghua University \quad
        $^3$Shanghai Jiao Tong University \\ \vspace{0.1cm}
        $^4$Zhejiang University \quad
        $^5$Nanyang Technological University \quad
        $^6$Harbin Institute of Technology, Shenzhen \\ \vspace{0.1cm}
        $^7$Institute of Automation, CASIA \quad
        $^8$Taiyuan University of Technology \quad
        $^9$Shanghai AI Laboratory
    }
}

\abstract{
While Multimodal Large Language Models (MLLMs) demonstrate proficiency in 2D scenes, extending their perceptual intelligence to 3D point cloud understanding remains a significant challenge. Current approaches focus primarily on aligning 3D features with pre-trained models. However, they typically treat geometric reasoning as an implicit mapping process. These methods bypass intermediate logical steps and consequently suffer from geometric hallucinations. They confidently generate plausible responses that fail to ground in precise structural details. To bridge this gap, we present PointCoT, a novel framework that empowers MLLMs with explicit Chain-of-Thought (CoT) reasoning for 3D data. We advocate for a \textit{Look, Think, then Answer} paradigm. In this approach, the model is supervised to generate geometry-grounded rationales before predicting final answers. To facilitate this, we construct Point-Reason-Instruct, a large-scale benchmark comprising $\sim$86k instruction-tuning samples with hierarchical CoT annotations. By leveraging a dual-stream multi-modal architecture, our method synergizes semantic appearance with geometric truth. Extensive experiments demonstrate that PointCoT achieves state-of-the-art performance on complex reasoning tasks.

\noindent \textbf{Keywords:} 3D Vision, Point Cloud Understanding, Multimodal Large Language Models, Chain-of-Thought, Multi-modal Learning}

\metadata[Contact]{Dongxu Zhang, \email{zhangdongxu@stu.xjtu.edu.cn} % 替换为你的邮箱
\\
\\
  \parbox{\linewidth}{\centering
    \github~\href{https://github.com/DongXu-Zhang/PointCot}{\textbf{Code}} \quad
    \web~\href{https://dongxu-zhang.github.io/PointCoT.github.io/\#}{\textbf{Website}} \quad
    \huggingface~\href{https://huggingface.co/datasets/dongxu0852/Point-Reason-Instruct}{\textbf{Dataset}}
  }
}

\begin{document}

\maketitle

% \tableofcontents

\section{Introduction}
\label{sec:int}
\makeatletter
\def\@makefnmark{}
\makeatother
\footnotetext{* Equal contribution.}
Recent advancements in Multimodal Large Language Models (MLLMs), such as GPT-4V~\cite{achiam2023gpt} and LLaVA~\cite{liu2023visual}, have achieved remarkable progress in visual perception and reasoning. By aligning visual encoders with powerful Large Language Models (LLMs)~\cite{lin2025hidden}, these agents demonstrate exceptional proficiency in perceiving and reasoning about 2D planar images. However, the physical world we inhabit is inherently three-dimensional. To build truly embodied agents, capable of navigating environments, manipulating objects, and interacting with the physical world, extending this perceptual intelligence from 2D pixels to 3D geometric structures (\eg, point clouds) is an imperative yet challenging frontier.

Driven by this necessity, pioneering works like Point-LLM~\cite{xu2024pointllm} and 3D-LLM~\cite{hong20233d} have successfully projected 3D point cloud features into the LLM input space, enabling basic 3D Question Answering. Despite these advancements, a critical limitation remains overlooked: most existing paradigms treat 3D reasoning as a black-box mapping process. These methods train models to map input point clouds directly to final answers end-to-end, bypassing explicit reasoning steps. 

\begin{figure*}[t]
  \centering
  \includegraphics[width=0.97\linewidth]{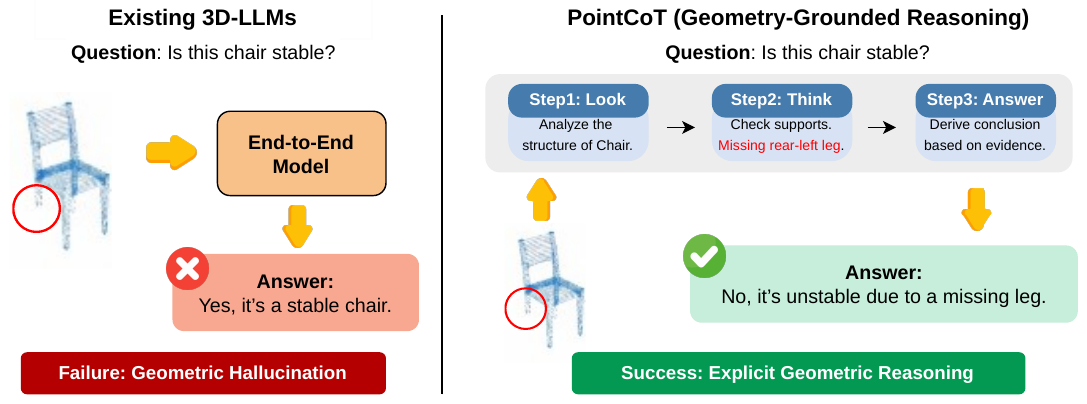}
  \caption{Existing 3D-LLMs (left) treat geometric reasoning as a direct mapping process, often suffering from Geometric Hallucination, failing to perceive the missing leg and incorrectly judging the chair as stable. In contrast, our proposed PointCoT (right) introduces an explicit Look-Think-Answer paradigm. By generating a geometry-grounded rationale (detecting the missing rear-left leg) before the final conclusion, our method significantly reduces hallucinations and enables interpretable 3D reasoning.}
  \label{fig:motivation}
\end{figure*}

However, when facing complex spatial tasks, such as judging whether a chair with a missing leg is stable, these models often suffer from what we term Geometric Hallucination~\cite{li2023evaluating,liu2023hallusionbench}. As illustrated in \cref{fig:motivation} (left), a conventional model might correctly identify the semantic category but fail to ground its judgment in fine-grained geometric details (\eg, the missing leg), leading to plausible but factually incorrect conclusions.
Without an explicit reasoning chain, the decision-making process of current 3D-LLMs remains opaque and unreliable.

Bridging this interpretability gap requires a fundamental shift in how we model 3D understanding. We draw inspiration from the Natural Language Processing (NLP), where Chain-of-Thought (CoT) prompting~\cite{wei2022chain,zhang2025ascot,zhang2026chain} has proven effective in eliciting complex reasoning by decomposing problems into intermediate steps. We argue that robust 3D understanding demands a similar \textit{Look, Think, then Answer} mechanism, as depicted in \cref{fig:motivation} (right). Specifically, a 3D agent should first perceive the fine-grained geometry (Look), derive an explicit rationale grounded in spatial evidence (Think), and finally deduce the conclusion (Answer). Nevertheless, transposing this CoT paradigm to the 3D domain is non-trivial, as it faces two substantial hurdles. The primary bottleneck is Data Scarcity, existing 3D benchmarks~\cite{azuma2022scanqa,ma2022sqa3d} offer simple pair-wise annotations, lacking the explicit rationale supervision required to train the reasoning process. This challenge is further compounded by the Modality Gap. Unlike rich 2D images, point clouds are sparse and lack semantic texture~\cite{qi2017pointnet++,xue2023ulip}, making pure geometric reasoning difficult, while rendered images often suffer from depth ambiguity.

To address these challenges, we present PointCoT, the first framework to integrate explicit CoT reasoning into 3D point cloud understanding. To democratize research in this direction, we first construct Point-Reason-Instruct, a large-scale object-level instruction-tuning dataset~\cite{guo2023point}. Unlike previous benchmarks, our dataset provides a triplet of $\langle$\textit{Point Cloud, Multi-view Images, CoT Rationale}$\rangle$, enabling models to learn how to reason, not just what to answer. Second, we propose a Multi-modal Synergistic Reasoning Framework. Recognizing that images offer rich semantics while point clouds provide geometric truth, we design a dual-stream encoder to fuse these modalities. We then employ a two-stage training paradigm to teach the LLM to generate geometry-grounded rationales before predicting the final answer. This explicit reasoning process significantly reduces hallucination and enhances interpretability.

Our main contributions are summarized as follows:
\begin{itemize}
    \item To the best of our knowledge, we are the first to transfer the explicit CoT reasoning paradigm to the domain of 3D point cloud understanding. We shift the 3D learning from implicit end-to-end mapping to a transparent \textit{Look-Think-Answer} mechanism, effectively mitigating geometric hallucinations.
    \item We construct \textit{Point-Reason-Instruct}, the first large-scale dataset combining 3D point clouds with explicit CoT annotations. It serves as a comprehensive benchmark for evaluating 3D reasoning capabilities.
    \item We propose \textit{PointCoT} that seamlessly integrates 3D point clouds, multi-view images, and textual instructions. By employing a dual-stream encoder, our method effectively synergizes the structural precision of geometric data with the rich semantics of visual appearance.
    \item Extensive experiments demonstrate that \textit{PointCoT} achieves state-of-the-art performance on complex 3D reasoning tasks. Furthermore, our method exhibits superior interpretability and strong generalization capabilities.
\end{itemize}

\section{Related Work}
\label{sec:rel}
\subsection{3D Representation Learning}
Learning robust representations from irregular point clouds is foundational to 3D understanding~\cite{li2025pointdico,sun2025hyperpoint}. 
The field has evolved from MLP-based architectures like PointNet++~\cite{qi2017pointnet,qi2017pointnet++} to Transformer-based paradigms that capture complex global dependencies. Notable methods methods, such as PointBERT~\cite{yu2022point} and Point-MAE~\cite{pang2023masked}, leverage masked modeling to pre-train potent 3D encoders. 
Advanced attention mechanisms have also proven effective in geometric matching tasks~\cite{zhang2026igasa} and scalable architectures~\cite{qian2022pointnext}.
In parallel, multi-view approaches~\cite{liang2022mvcnn,goyal2021revisiting} project 3D shapes to leverage 2D priors. 
However, these methods typically yield static geometric abstractions limited to discriminative tasks (\eg, classification). They lack the semantic depth and language interface required for open-ended reasoning. In contrast, our work bridges this gap by integrating geometric features with explicit CoT reasoning, enabling the model to not only perceive 3D structures but also interpret them logically.

\subsection{Multi-modal Large Language Models}
The remarkable success of MLLMs in the 2D domain~\cite{liu2023visual,achiam2023gpt,lin2025orthalign} has propelled the extension of this paradigm to 3D understanding. 
Early alignment-centric works, such as ULIP~\cite{xue2023ulip} and Point-Bind~\cite{guo2023point}, focused on establishing joint embedding spaces to bridge 3D geometries with linguistic semantics. 
Building on this foundation, generative 3D-LLMs~\cite{hong20233d,xu2024pointllm} have emerged, injecting projected 3D features into LLMs to facilitate open-ended instruction following. 
However, existing 3D-LLMs primarily concentrate on feature alignment and learn a direct, end-to-end mapping from 3D inputs to answers. They treat the reasoning process as a black box, skipping the intermediate logical steps required for complex decision-making, rendering them prone to geometric hallucinations~\cite{li2023evaluating}, especially in tasks demanding rigorous spatial deduction. 
In contrast, we depart from this implicit paradigm by introducing explicit CoT reasoning, ensuring responses are derived from transparent, geometry-grounded logical paths.

\subsection{Chain-of-Thought Reasoning}
Originating from NLP, CoT prompting~\cite{wei2022chain,kojima2022large} has revolutionized LLM reasoning by decomposing complex problems into intermediate logical steps. 
This paradigm has been successfully adapted to the 2D visual domain, where frameworks like Multimodal-CoT~\cite{zhang2023multimodal,lu2022learn} achieve significant gains by generating textual rationales. 
Similarly, general-purpose VLMs~\cite{liu2023visual,bai2025qwen2,dai2023instructblip} acquire implicit reasoning capabilities through large-scale instruction tuning. 
However, the application of explicit CoT in 3D point cloud understanding remains largely unexplored. 
While recent works utilize proprietary models (\eg, GPT-4) for auxiliary 3D reasoning~\cite{sun20253d,yang2024llm}, they rely on external APIs and lack an end-to-end framework for learning internal geometric representations. 
We bridge this gap by integrating explicit, step-by-step geometric reasoning directly into the training pipeline, enabling models to derive answers from grounded 3D rationales.

\section{The Point-Reason-Instruct Benchmark}
\label{sec:ben}
Current 3D question-answering benchmarks predominantly rely on direct input-output pairs (\eg, ScanQA~\cite{azuma2022scanqa}), ignoring the intermediate reasoning chains essential for deep geometric understanding. This absence of explicit rationale supervision impedes the development of interpretable and robust 3D agents. To bridge this critical gap, we introduce Point-Reason-Instruct, a large-scale and geometrically grounded instruction-tuning benchmark designed to teach models how to think. In this section, we detail the multi-modal data generation, the scalable annotation pipeline, and the rigorous evaluation protocols.

\subsection{Data Sourcing and Multi-View Geometry Generation}
To ensure both geometric diversity and semantic richness, we curate our dataset from the Objaverse-LVIS~\cite{deitke2023objaverse} subset. We implement a topology-aware filtering protocol, prioritizing objects with complex geometries and articulated parts (\eg, furniture with multiple components) while filtering out overly simplistic shapes, ensuring the data supports intricate reasoning tasks.

\noindent\textbf{\textit{Geometric Normalization.}}
For each object, we perform canonical preprocessing to ensure spatial consistency. The raw meshes are centered at the origin and normalized to fit within a unit sphere. 
We then generate the point cloud input $\mathcal{P}$ by sampling $N=8,192$ points using Farthest Point Sampling (FPS) on the mesh surface. This density is empirically chosen to balance the preservation of fine-grained structural details with computational efficiency.

\noindent\textbf{\textit{Holistic Multi-View Rendering.}}
While point clouds provide precise coordinates, they inherently lack dense semantic textures. To complement this, we generate a set of auxiliary multi-view images $\mathcal{I}$. 
Existing protocols often rely solely on horizontal viewpoints, leaving critical areas blind. To address this, we design a Spherical 8-View System for holistic coverage. 
Specifically, we render RGB images from six equidistant azimuth angles ($0^\circ$ to $300^\circ$) at a $30^\circ$ elevation. 
Crucially, these are complemented by two orthogonal views (zenith and nadir) to capture cavities, inner surfaces, and supporting mechanisms (\eg, chassis) that are typically occluded in canonical horizontal scans.
This omnidirectional design allows the model to reason about object affordances (\eg, containability) that require visibility of the object's interior or underside.

\begin{figure*}[t]
  \centering
  \includegraphics[width=0.97\linewidth]{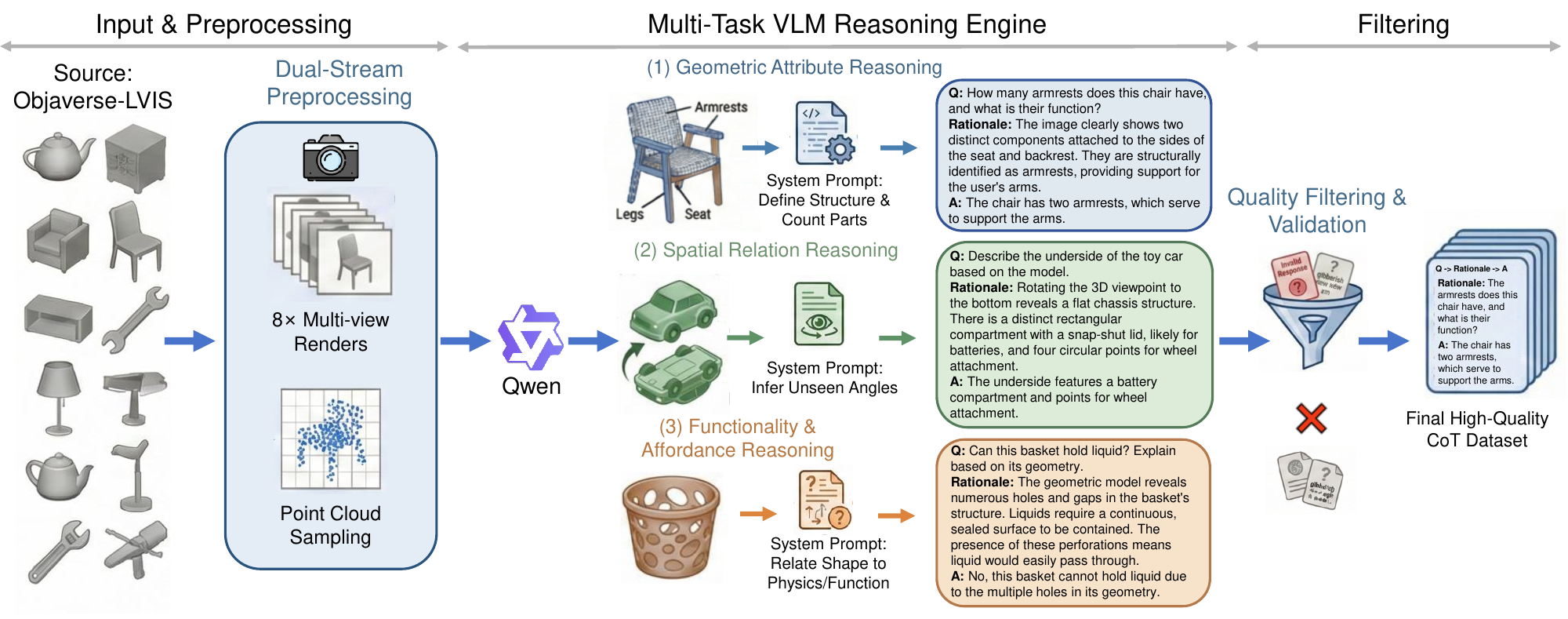}
  \caption{The Data Construction Pipeline of Point-Reason-Instruct. The pipeline consists of three stages: (1) Dual-Stream Preprocessing, where objects are sampled into point clouds and rendered into 8 spherical views; (2) Multi-Task Reasoning Generation, where the Qwen2.5-VL teacher agent generates hierarchical CoT rationales covering geometric attributes, spatial relations, and functionality; and (3) Quality Filtering, which validates the rationales against metadata to eliminate hallucinations.}
  \label{fig:dataset}
\end{figure*}

\subsection{Scalable Annotation with Qwen2.5-VL}
Manually annotating step-by-step geometric reasoning for $\sim$86k samples is computationally intractable. To address this, we construct an automated Geometry-Aware Teacher Agent pipeline driven by Qwen2.5-VL-72B-Instruct~\cite{bai2025qwen2}. We select this model for its state-of-the-art proficiency in high-resolution visual grounding, which is imperative for perceiving the fine-grained spatial details.

\noindent\textbf{\textit{Rationale Verification and Filtering.}}
While 2D VLMs provide strong semantic priors, they are inherently prone to spatial hallucinations. To ensure 3D topological fidelity, we decouple rationale generation from geometric verification. We utilize the VLM solely as a candidate proposer and enforce a deterministic cross-validation protocol. Specifically, every spatial assertion parsed from the generated CoT is rigorously cross-referenced against the rigid 3D object metadata and multi-view consistency constraints. Candidates failing this explicit geometric verification are discarded, guaranteeing that the final training corpus is anchored in physical 3D truth rather than biased 2D visual priors.

\subsection{Hierarchical Task Design}
To rigorously evaluate reasoning depth, we structure the Point-Reason-Instruct dataset into three cognitive hierarchies, ranging from local perception to abstract deduction. The dataset accumulates a total of 86,280 instruction-tuning samples.

\noindent\textbf{\textit{Level 1: Structural Part Reasoning ($\sim$28k).}}
The foundational tier focuses on the identification and structural analysis of explicit object components. Unlike simple attribute classification, queries here require the model to recognize specific parts (\eg, armrests, legs, chassis), count them, and analyze their intrinsic geometric integrity.
\textit{Example: ``How many armrests does this chair have, and are they connected to a central axis?''}

\noindent\textbf{\textit{Level 2: 3D Viewpoint Reasoning ($\sim$29k).}}
Ascending in complexity, this subset targets the holistic understanding of 3D structures and spatial perspectives. The model is challenged to perform mental rotation or infer details of occluded viewpoints (\eg, the underside or rear surface) based on the visible geometry, as well as reasoning about the relative spatial arrangement.
\textit{Example: ``Describe the geometric structure of the chair's back from the rear view.''}

\noindent\textbf{\textit{Level 3: Functionality and Affordance Reasoning ($\sim$29k).}}
The most advanced tier requires physics-grounded causal reasoning. Here, the model must bridge the gap between static form and dynamic function, applying physical principles (\eg, gravity, friction, and containment) to deduce potential interactions or usage scenarios directly from the geometry.
\textit{Example: ``Would this container spill its contents if tilted 45 degrees?''}

\subsection{Strict Object-Level Splitting}
A pervasive issue in existing 3D learning benchmarks is data leakage~\cite{uy2019revisiting}, where different views or questions about the same object appear in both training and testing sets. To ensure a rigorous evaluation of generalization rather than memorization, we implement a Strict Object-Level Splitting strategy.
This results in a final partition of 22,871 objects (79.5\%) for Training, 2,945 objects (10.2\%) for Validation, and 2,944 objects (10.2\%) for Testing. This guarantees that the geometric shapes encountered during inference are entirely unseen, providing a true test of the model's ability to reason about novel 3D structures.

\section{Method}
\label{sec:met}

\subsection{Overview and Problem Formulation}

While CoT prompting excels in NLP and 2D vision-language tasks, its extension to 3D point clouds is hindered by the sparse and geometrically irregular nature of 3D data. 
To transcend these limitations, we propose PointCoT, a transparent \textit{Look-Think-Answer} deductive paradigm. Instead of learning a black-box mapping to maximize the unconstrained likelihood $P(\mathcal{A} | \mathcal{P}, \mathcal{I}, \mathcal{Q})$, we reformulate 3D reasoning as a sequential decoding process. Given a raw point cloud $\mathcal{P}$, auxiliary multi-view images $\mathcal{I}$, and a linguistic instruction $\mathcal{Q}$, we introduce an explicit intermediate rationale $\mathcal{R}$ to formulate the optimal predicted answer $\hat{\mathcal{A}}$ by maximizing the expected probability over the answer space $\mathcal{A}$:
\begin{equation}
    \hat{\mathcal{A}} = \arg\max_{\mathcal{A}} \, \mathbb{E}_{\mathcal{R} \sim P_{\text{gen}}(\cdot | \mathbf{z})} \big[ P_{\text{pred}}(\mathcal{A} | \mathbf{z}, \mathcal{R}) \big], \quad \text{s.t.} \quad \mathbf{z} = \mathcal{E}_{\text{align}}(\mathcal{P}, \mathcal{I}, \mathcal{Q}).
\end{equation}

This probabilistic formulation operationalizes our paradigm through a tripartite architectural progression. During the \textit{Look} stage, the alignment function $\mathcal{E}_{\text{align}}$ projects the multi-modal inputs into a unified geometric-semantic manifold $\mathbf{z}$. Subsequently, in the \textit{Think} stage, $P_{\text{gen}}$ autoregressively samples the structural rationale $\mathcal{R}$, which is strictly bounded by the topological priors of $\mathbf{z}$. Finally, the \textit{Answer} stage employs $P_{\text{pred}}$ to deduce the final response $\mathcal{A}$ via a progressive dual-stage optimization.

\subsection{Tri-Modal Contextualization}

During the \textit{Look} stage, the primary objective of the alignment function $\mathcal{E}_{\text{align}}$ is to integrate the continuous 3D spatial domain, the discrete 2D projective plane, and the sequential 1D linguistic space into a unified target manifold $\mathbf{z}$.

\noindent{\textbf{\textit{Non-Verbal Sensory Perception.}}}
We first extract high-dimensional geometric representations $H_{geo} \in \mathbb{R}^{N_p \times D}$ and their absolute 3D spatial centroids $\mathcal{C}_{3D} \in \mathbb{R}^{N_p \times 3}$ from the raw point cloud $\mathcal{P}$ using a Point Encoder. Concurrently, a Vision Transformer embeds the multi-view observations $\mathcal{I}$ into a visual semantic space $H_{vis} \in \mathbb{R}^{N_v \times D}$, parameterized by 2D patch coordinates $\mathcal{C}_{2D} \in \mathbb{R}^{N_v \times 2}$.

\begin{figure*}[t]
  \centering
  \includegraphics[width=0.98\linewidth]{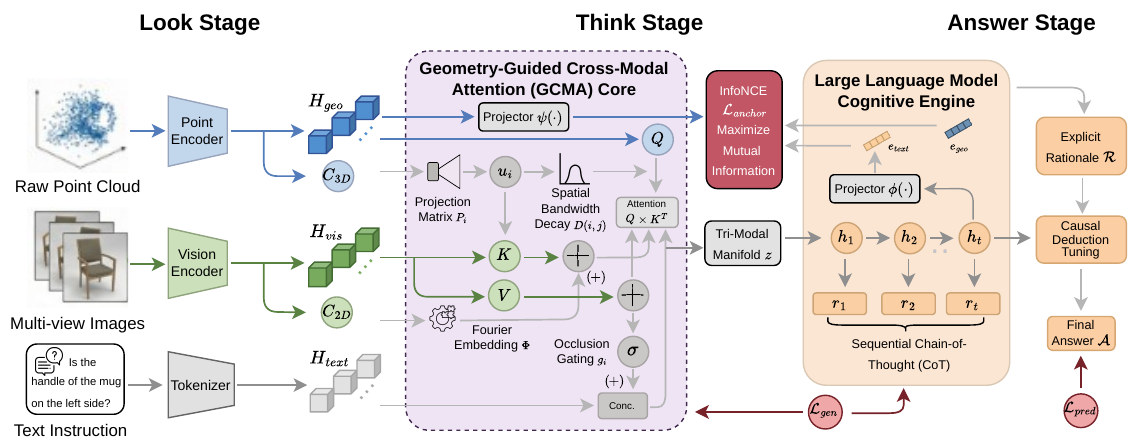}
  \caption{The overall architecture of PointCoT. The framework operates in a \textit{Look-Think-Answer} paradigm. In the Look Stage, a dual-stream encoder extracts geometric and visual features, which are fused into a Tri-Modal Manifold $\mathbf{z}$ via Geometry-Guided Cross-Modal Attention. During the Think Stage, a VLM autoregressively generates an explicit rationale $\mathcal{R}$, while its hidden states $h_t$ are strictly grounded to $H_{geo}$ via an InfoNCE loss $\mathcal{L}_{anchor}$ to mitigate spatial hallucinations. Finally, in the Answer Stage, the final answer $\mathcal{A}$ is deduced conditionally on both $\mathbf{z}$ and $\mathcal{R}$. The entire pipeline is trained through a progressive dual-stage optimization using $\mathcal{L}_{gen}$ and $\mathcal{L}_{pred}$.}
  \label{fig:pipeline}
\end{figure*}

\noindent{\textbf{\textit{Geometry-Guided Cross-Modal Attention.}}}
Standard cross-attention mechanisms treat sensory tokens as unstructured sets, omitting intrinsic 3D-to-2D homeomorphic projections. To address this, our Geometry-Guided Cross-Modal Attention (GCMA) module synchronizes $H_{geo}$ and $H_{vis}$ by modulating their interaction with physical projection priors. Let $Q_i = H_{geo}^{(i)} W_q$ be the query of the $i$-th 3D token, and $K_j = H_{vis}^{(j)} W_k$, $V_j = H_{vis}^{(j)} W_v$ be the key and value of the $j$-th 2D visual patch. Let $v(j)$ denote the camera view index corresponding to the $j$-th 2D visual patch. We map the 3D coordinate $p_i \in \mathcal{C}_{3D}$ onto the specific 2D projective plane of this view via its corresponding camera projection matrix $\Pi_{v(j)}$, yielding the view-dependent projected coordinate $u_{i,j} = \pi(p_i, \Pi_{v(j)})$. A localized spatial bandwidth constraint $\mathcal{D}(i, j)$ guides the cross-attention via isotropic Gaussian decay over the Euclidean distance between $u_{i,j}$ and the visual patch centroid $c_j \in \mathcal{C}_{2D}$:
\begin{equation}
    \mathcal{D}(i, j) = \exp\left( - \frac{\lVert u_{i,j} - c_j \rVert_2^2}{2\sigma_s^2} \right),
\end{equation}
where $\sigma_s$ is a learnable spatial bandwidth parameter.
To capture high-frequency morphological variations (\eg, sharp structural discontinuities), we concurrently incorporate a continuous Fourier-based relative spatial embedding $\Phi(p_i, c_j) = \mathcal{F}_\theta(\gamma(p_i) \oplus \gamma(c_j))$, where $\gamma(\cdot)$ is the spectral mapping and $\mathcal{F}_\theta$ is a non-linear network. Consequently, the composite attention logit $\mathbf{A}_{i,j}$ explicitly fuses semantic affinity, projection-guided localization, and geometric bias:
\begin{equation}
    \mathbf{A}_{i,j} = \left( \frac{Q_i K_j^\top}{\sqrt{d}} \right) \cdot \mathcal{D}(i, j) + \Phi(p_i, c_j),
\end{equation}
where $d$ is the feature dimension of the queries and keys.

\noindent{\textbf{\textit{Tri-Modal Sequence Formulation.}}}
To mitigate the occlusion from 3D-to-2D projections, we regulate the semantic feature $\hat{H}_{geo}^{(i)} = \sum_{j} \text{softmax}_j (\mathbf{A}_{i,j}) V_j$ with a dynamic occlusion-aware gate $g_i = \sigma ( \mathcal{T}_\phi ( H_{geo}^{(i)}, \hat{H}_{geo}^{(i)} ) )$. 
This produces a unified embedding token $H_{\text{sensory}}^{(i)} = \Psi_{\text{norm}} \big( H_{geo}^{(i)} + g_i \cdot \hat{H}_{geo}^{(i)} \big)$. Let $H_{\text{sensory}}$ denote the collection of all such tokens. Finally, we map $H_{\text{sensory}}$ into the LLM's semantic latent space and concatenate it with the tokenized linguistic instruction $H_{\text{text}}$, forming the fused target manifold $\mathbf{z} = [H_{\text{sensory}} \mathbin{\|} H_{\text{text}}]$ that conditions the subsequent \textit{Think} stage.

\subsection{Explicit Geometry-Grounded CoT Generation}
In the \textit{Think} stage, we strictly enforce structural verification prior to deductive resolution. The fused geometric-semantic manifold $\mathbf{z}$ serves as the conditional prefix for the LLM. The cognitive engine is trained to autoregressively unfold a discrete Markov chain of reasoning tokens $\mathcal{R} = \{r_t\}_{t=1}^T$. During inference, this decoding trajectory is formulated as a maximum a posteriori estimation, optimizing the log-likelihood of the logical sequence:
\begin{equation}
    \hat{\mathcal{R}} = \arg\max_{\mathcal{R}} \sum_{t=1}^{T} \log P_{\text{gen}}(r_t | r_{<t}, \mathbf{z}).
\end{equation}

\noindent{\textbf{\textit{Mitigation of Spatial Hallucinations.}}}
Multi-modal networks frequently suffer from spatial hallucinations by overfitting to 2D semantic priors while disregarding 3D structural realities. To structurally neutralize this reliance on spurious correlations, we establish a geometric attention anchor using high-fidelity rationale supervision from the Point-Reason-Instruct dataset. At each decoding step $t$, linguistic concepts are recursively grounded back to the invariant geometric tokens $H_{geo}$. We formalize this by maximizing the mutual information between the evolving reasoning hidden state $h_t$ and the point cloud manifold via a contrastive InfoNCE objective:
\begin{equation}
    \mathcal{L}_{\text{anchor}} = - \frac{1}{T} \sum_{t=1}^{T} \log \frac{\exp\big( \text{sim}(\phi(h_t), \psi(H_{geo}^+)) / \tau \big)}{\sum_{k=1}^{K} \exp\big( \text{sim}(\phi(h_t), \psi(H_{geo}^{(k)})) / \tau \big)},
\end{equation}
where $\phi(\cdot)$ and $\psi(\cdot)$ denote continuous non-linear projections into a shared metric space, $\tau$ is the temperature scaling factor, 
$H_{geo}^+$ represents the global geometry embedding of the matched 3D instance, and $\{H_{geo}^{(k)}\}_{k=1}^{K}$ are the in-batch negative instance geometries. Applied densely at each decoding step $t$, this instance-level objective implicitly drives fine-grained token-level alignment.
% and $H_{geo}^+$ along with $\{H_{geo}^{(k)}\}_{k=1}^{K}$ represent the matched positive geometry and in-batch negative samples, respectively. 
Minimizing this intrinsically maximizes a variational lower bound on the conditional mutual information $I(\mathcal{R}; H_{geo} | H_{vis})$. This theoretical guarantee ensures that the generated logical sequence $\mathcal{R}$ is conditioned on physical 3D truths.

\subsection{Progressive Dual-Stage Optimization}

To circumvent the pitfalls of direct empirical risk minimization, we formulate a dual-stage curriculum optimization.
This strategy minimizes a decoupled composite objective, defined by the expected negative log-likelihoods of both rationale generation and answer prediction, heavily regularized by the geometric anchor loss, supervised by the ground-truth data distribution $\mathcal{D}$:
\begin{equation}
    \mathcal{L}_{\text{total}} = \mathbb{E}_{(\mathbf{z}, \mathcal{R}^*, \mathcal{A}^*) \sim \mathcal{D}} \Big[ -\log P_{\text{gen}}(\mathcal{R}^* | \mathbf{z}) - \lambda \log P_{\text{pred}}(\mathcal{A}^* | \mathbf{z}, \mathcal{R}^*) \Big] + \alpha \mathcal{L}_{\text{anchor}},
    \label{eq:loss}
\end{equation}
where $\lambda$ and $\alpha$ are balancing coefficients.

\noindent{\textbf{\textit{Stage I: Reasoning Initialization.}}}
We first align the cross-modal manifold $\mathbf{z}$ with the LLM's inferential space by jointly optimizing the rationale likelihood and the geometric grounding objective: $\mathcal{L}_{\text{stage1}} = \mathbb{E}_{(\mathbf{z}, \mathcal{R}^*) \sim \mathcal{D}} [-\log P_{\text{gen}}(\mathcal{R}^* | \mathbf{z})] + \alpha \mathcal{L}_{\text{anchor}}$. To mathematically isolate the \textit{Think} stage, we enforce strict gradient truncation on the answer prediction objective ($\mathcal{L}_{\text{pred}} \rightarrow 0$). This forces the acquisition of explicit geometric reasoning by preventing the network from exploiting latent dataset biases to shortcut the final answer.

\noindent{\textbf{\textit{Stage II: Causal Deduction Tuning.}}}
We subsequently advance to the joint objective $\mathcal{L}_{\text{total}}$. Instead of relying on discrete sampling during training, we employ standard teacher forcing. By providing the ground-truth explicit reasoning trajectory $\mathcal{R}^*$ as a contextual prefix, we optimize the network to deduce the final answer $\mathcal{A}^*$. 
While teacher forcing risks exposure bias during inference, our $\mathcal{L}_{\text{anchor}}$ strictly bounds the autoregressive search space to $\mathbf{z}$. This geometric grounding intrinsically acts as a self-correcting mechanism, preventing severe error cascading along the generated sequence $\hat{\mathcal{R}}$ to ensure robust deductive resolution.

\begin{table}[t]
\centering
\caption{Quantitative results on the Point-Reason-Instruct benchmark. We report the Accuracy (\%) across three cognitive dimensions. Geo. evaluates fundamental structural perception; Spat. assesses relative positioning and connectivity; Func. measures high-level deductive reasoning regarding physical utility.}
\label{tab:main_results}
\resizebox{\textwidth}{!}{
\begin{tabular}{l|c|l|c|ccc}
\toprule
\textbf{Model} & \textbf{Modality} & \textbf{Backbone (Encoder / LLM)} & \textbf{Overall} & \textbf{Geo.} & \textbf{Spat.} & \textbf{Func.} \\
\midrule
\multicolumn{7}{l}{\textit{General-Purpose 2D VLMs (Zero-shot*)}} \\
GPT-4V \cite{achiam2023gpt} & Img & Proprietary / GPT-4 & 65.4 & 58.2 & 68.5 & 71.2 \\
Qwen2-VL-7B \cite{bai2025qwen2} & Img & ViT-L / Qwen2 & 63.8 & 54.5 & 64.2 & 69.5 \\
LLaVA-1.5 \cite{liu2023visual} & Img & ViT-L / Vicuna-7B & 54.1 & 47.2 & 51.8 & 58.9 \\
\midrule
\multicolumn{7}{l}{\textit{Specialized 3D-LLMs (Fine-tuned)}} \\
Chat-3D v2 \cite{huang2023chat} & PC & ViT-g / Vicuna-7B & 66.1 & 72.4 & 63.2 & 62.8 \\
Point-LLM \cite{xu2024pointllm} & PC & PointBERT / Vicuna-7B & 62.4 & 68.1 & 59.2 & 58.5 \\
Point-Bind \cite{guo2023point} & PC & PointBERT / Llama2-7B & 58.1 & 65.2 & 55.4 & 52.3 \\
\midrule
\textbf{PointCoT (Ours)} & \textbf{PC+Img} & \textbf{PointBERT / Qwen2.5-7B} & \textbf{78.5} & \textbf{82.3} & \textbf{76.4} & \textbf{75.1} \\
\bottomrule
\end{tabular}}
\end{table}

\section{Experiments}
\label{sec:exp}

\subsection{Experimental Setup}

\noindent{\textbf{\textit{Datasets.}}}
We primarily evaluate PointCoT on our proposed Point-Reason-Instruct benchmark, comprising $\sim$86k instruction-tuning samples across three hierarchical subsets. To assess zero-shot generalization, we additionally employ Objaverse-LVIS~\cite{deitke2023objaverse} for open-vocabulary 3D captioning and ScanQA~\cite{azuma2022scanqa} for complex indoor spatial reasoning.

\noindent{\textbf{\textit{Implementation Details.}}}
Our architecture integrates PointBERT~\cite{yu2022point} and EVA-CLIP (ViT-g/14)~\cite{dosovitskiy2020image} to encode 3D geometries and 2D semantics, with Qwen2.5-7B-Instruct~\cite{bai2025qwen2} serving as the cognitive LLM backbone. Training proceeds in two stages. The initial \textit{CoT Generation Tuning} stage aligns multi-modal features with the LLM to elicit explicit rationales. A subsequent \textit{End-to-End Tuning} phase then optimizes final answer prediction conditioned on the generated reasoning chains. Models are trained on eight NVIDIA A100 (80GB) GPUs. For a fair comparison, all 3D-LLM baselines are fine-tuned on our dataset. Detailed hyperparameters, including learning rates and loss balancing coefficients ($\lambda, \alpha$), are provided in the Supplementary Material.

\noindent{\textbf{\textit{Evaluation Metrics.}}}
We adopt a hierarchical evaluation protocol. General task performance is measured via Exact Match Accuracy for classification tasks and BLEU-4 for open-ended generation benchmarks. To evaluate the faithfulness of intermediate rationales, we introduce two specialized metrics. The first is a GPT-4 Reasoning Score ranging from 1 to 10 for assessing logical coherence and factual correctness. The second is the Geometric Hallucination Rate (GHR), which quantifies the percentage of generated spatial assertions contradicting ground-truth metadata to serve as a strict proxy for geometric grounding.

\subsection{Main Results}

\noindent{\textbf{\textit{Comparison with State-of-the-Art.}}}
Table~\ref{tab:main_results} demonstrates that PointCoT establishes a new state-of-the-art on the Point-Reason-Instruct benchmark with an overall accuracy of \textbf{78.5\%}. The results expose critical failure modes in existing paradigms. First, general-purpose 2D VLMs suffer severe performance drops on Geo. (58.2\% vs. our 82.3\%), confirming an inherent discrepancy: purely projective visual representations lack the explicit 3D coordinate priors necessary for perceiving intrinsic structural features.  Second, conventional 3D baselines falter on Func. ($\sim$60\%), indicating that implicit black-box mappings fail to elicit the physical commonsense and causal reasoning required for high-level tasks. In contrast, PointCoT achieves balanced superiority across all subsets. By synergizing geometric precision with semantic logic, our method outperforms the strongest fine-tuned baseline by \textbf{+12.4\%}, firmly validating the efficacy of the explicit \textit{Look-Think-Answer} paradigm in mitigating geometric hallucinations.

\begin{table}[t]
\centering
\caption{Evaluation of architectural agnosticism. We report the Overall Accuracy (\%) and GPT-4 based Reasoning Score (1-10) for PointCoT instantiated with varying LLM backbones and 3D point encoders. The results demonstrate steady scalability and robustness across diverse network configurations. Best results are highlighted in bold.}
\label{tab:robustness}
\setlength{\tabcolsep}{10pt}
\resizebox{0.98\textwidth}{!}{
\begin{tabular}{l|c|c|c}
\toprule
\textbf{LLM Backbone} & \textbf{Point Encoder} & \textbf{Overall Acc (\%)} & \textbf{Reasoning Score} \\
\midrule
Vicuna-7B & PointBERT & 73.4 & 7.9 \\
Mistral-7B-v0.3 & PointBERT & 76.8 & 8.3 \\
Qwen2.5-7B (Ours) & PointBERT & 78.5 & 8.5 \\
Qwen2.5-7B & PointNeXt & \textbf{79.6} & \textbf{8.7} \\
\bottomrule
\end{tabular}}
\end{table}

\begin{table}[t]
\centering
\caption{Zero-Shot Transfer Evaluation. We assess generalization capability on unseen benchmarks. ScanQA evaluates scene-level spatial reasoning (BLEU-4), while Objaverse evaluates open-vocabulary classification (Accuracy). $^\dagger$ indicates evaluation on localized instance crops to align with our object-centric training regime.}
\label{tab:zeroshot}
\setlength{\tabcolsep}{8pt}
\resizebox{1.0\textwidth}{!}{
\begin{tabular}{lcccc}
\toprule
\textbf{Method} & \textbf{Training Data (Source)} & \textbf{Data Scale} & \textbf{ScanQA$^\dagger$} & \textbf{Objaverse} \\
\midrule
3D-LLM~\cite{hong20233d} & 300k (3D-Data) & 4.3x & \textbf{24.5} & 49.2 \\
Point-LLM~\cite{xu2024pointllm} & 660k (Cap3D) & 9.6x & 22.4 & 45.1 \\
\midrule
\textbf{PointCoT (Ours)} & \textbf{$\sim$69k (Ours)} & \textbf{1.0x} & 23.4 & \textbf{51.8} {\scriptsize (+2.6)} \\
\bottomrule
\end{tabular}}
\end{table}

\noindent{\textbf{\textit{Architecture Agnosticism and Scalability.}}}
To verify that performance gains stem from the explicit \textit{Look-Think-Answer} paradigm rather than mere LLM capacity, we evaluate PointCoT across diverse architectures. As shown in Table~\ref{tab:robustness}, pairing PointCoT with a weaker backbone like Vicuna-7B yields 73.4\% overall accuracy. This significantly outperforms the end-to-end Point-LLM baseline (62.4\% in Table 1), confirming the intrinsic value of explicit rationale generation. Furthermore, upgrading to Mistral-7B and Qwen2.5-7B provides predictable scalability, reaching 76.8\% and 78.5\%, respectively. Finally, substituting the geometric frontend with PointNeXt pushes the accuracy to \textbf{79.6\%} , validating PointCoT as a versatile, architecture-agnostic framework for robust 3D reasoning.

\noindent{\textbf{\textit{Zero-Shot Generalization and Data Efficiency.}}}
To assess transferable physical intelligence, we evaluate zero-shot performance on ScanQA~\cite{azuma2022scanqa} and Objaverse-LVIS~\cite{deitke2023objaverse}. 
We conduct an evaluation centered on individual objects for the ScanQA benchmark because PointCoT is trained on local instance geometries rather than complex room layouts. We deliberately factor out global relational parsing by isolating target instances utilizing ground truth bounding boxes.
% To bridge the gap between our object-centric training and ScanQA's multi-object scenes, we isolate target instances using ground-truth bounding boxes for localized geometric reasoning. 
As Table~\ref{tab:zeroshot} shows, PointCoT demonstrates exceptional data efficiency. Despite training on merely $\sim$69k samples, it achieves a leading 51.8\% accuracy on Objaverse-LVIS. On ScanQA, while predictably trailing the scene-pretrained 3D-LLM, PointCoT surpasses its object-centric counterpart Point-LLM. By mastering localized geometric primitives (\eg, curvature, topology), PointCoT maximizes zero-shot generalization across domain shifts under severe data constraints.

\noindent{\textbf{\textit{Quantitative Assessment of Reasoning Quality.}}}
To objectively evaluate reasoning fidelity, we employ GPT-4 as an expert judge, conditioning it on ground-truth 3D metadata to prevent language-prior bias. We evaluate 200 sampled rationales from GPT-4V, a CoT-adapted Point-LLM, and PointCoT. As shown in Table~\ref{tab:reasoning_score}, PointCoT consistently outperforms all baselines. Most notably, it achieves a decisive advantage in the \textit{Grounding} metric. This significant margin confirms that our deductive chains are firmly anchored in verifiable 3D spatial evidence rather than plausible textual hallucinations, effectively bridging the interpretability gap between perception and generation.

\begin{table}[t]
\centering
\caption{Automated Evaluation of Rationale Quality via GPT-4. Scores range from 1 to 10. The evaluator is strictly conditioned on ground-truth 3D metadata to ensure objective geometric verification.}
\label{tab:reasoning_score}
\setlength{\tabcolsep}{10pt}
\resizebox{1.0\textwidth}{!}{
\begin{tabular}{lccc|c}
\toprule
\textbf{Method} & \textbf{Correctness} & \textbf{Logic} & \textbf{Grounding} & \textbf{Average} \\
\midrule
Point-LLM (CoT-adapted) & 7.1 & 6.8 & 6.2 & 6.7 \\
GPT-4V & 8.1 & 8.4 & 7.6 & 8.0 \\
\midrule
\textbf{PointCoT (Ours)} & \textbf{8.4} & \textbf{8.6} & \textbf{8.9} & \textbf{8.6} \\
\bottomrule
\end{tabular}}
\end{table}

\subsection{Qualitative Analysis}
\label{sec:qualitative}

\begin{table}[t]
\centering
\caption{Ablation of input modalities and reasoning strategies. (a) Hybrid modality synergizes semantic and geometric precision. (b) Explicit CoT significantly suppresses geometric hallucinations compared to black-box mapping.}
\label{tab:ablation_all}
\begin{subtable}[t]{0.48\textwidth}
\centering
\begin{tabular}{lcc}
\toprule
\textbf{Configuration} & \textbf{Acc (\%)} & \textbf{Score} \\
\midrule
Image Only & 55.2 & 5.1 \\
Point Only & 64.8 & 5.8 \\
\textbf{Hybrid (Ours)} & \textbf{78.5} & \textbf{8.5} \\
\bottomrule
\end{tabular}
\caption{Input Modalities}
\end{subtable}
\hfill
\begin{subtable}[t]{0.48\textwidth}
\centering
\begin{tabular}{lcc}
\toprule
\textbf{Strategy} & \textbf{Acc (\%)} & \textbf{GHR} \\
\midrule
Direct Map & 67.4 & 25.4\% \\
Implicit CoT & 71.2 & 18.2\% \\
\textbf{Explicit CoT} & \textbf{78.5} & \textbf{5.1\%} \\
\bottomrule
\end{tabular}
\caption{Reasoning Strategy}
\end{subtable}
\end{table}

\begin{figure*}[t]
  \centering
  \includegraphics[width=0.97\linewidth]{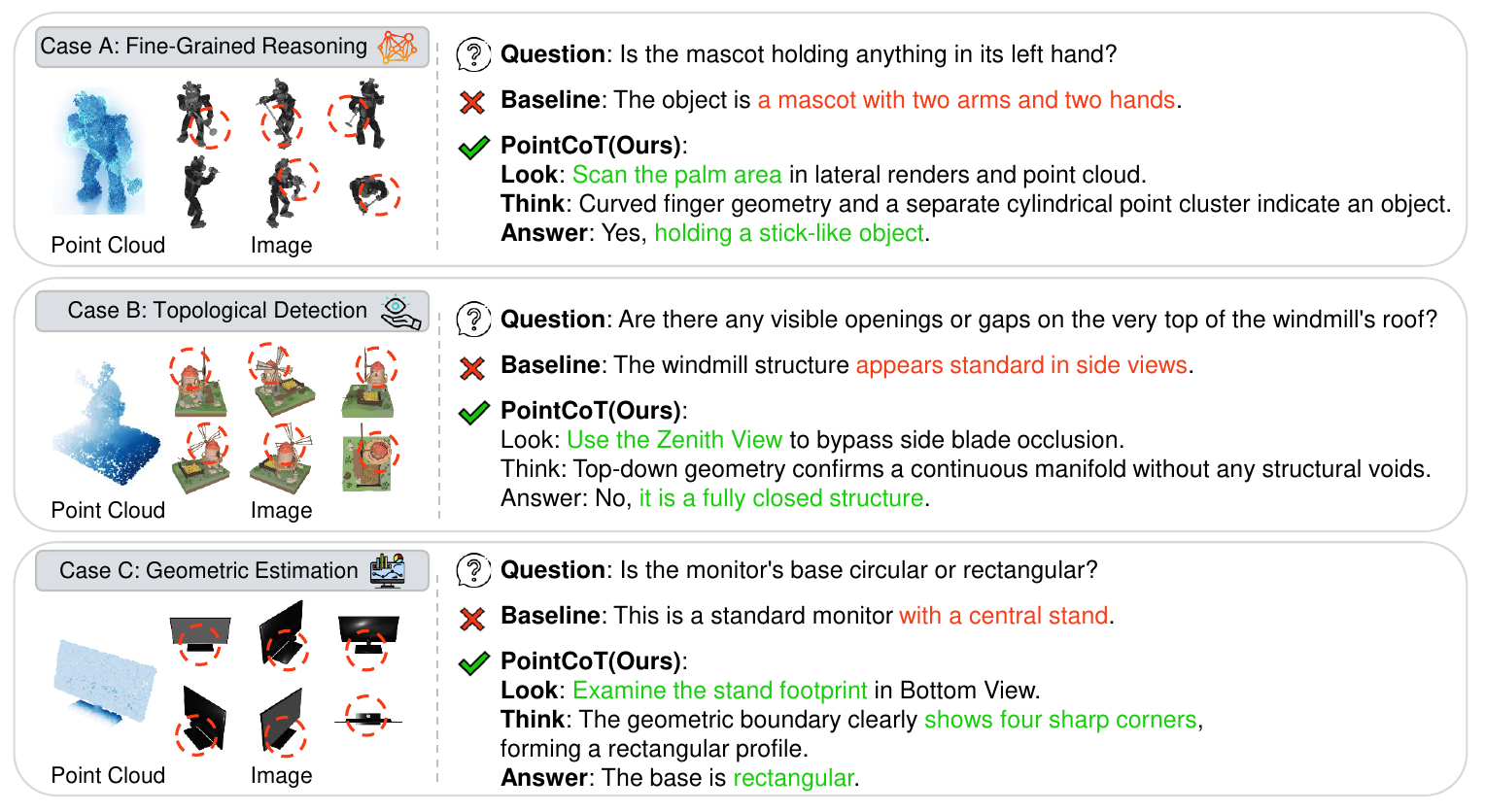}
  \caption{Qualitative comparison on the Point-Reason-Instruct benchmark. While the baseline relies on implicit semantic priors and suffers from geometric hallucinations, PointCoT follows a \textit{Look-Think-Answer} paradigm. By actively selecting decisive viewpoints and explicitly grounding rationales in local 3D geometric structures, PointCoT effectively mitigates hallucinations and yields interpretable predictions.}
  \label{fig:qualitative}
\end{figure*}

\noindent{\textbf{\textit{Visualizations and Case Studies.}}}
We provide qualitative comparisons in \cref{fig:qualitative} to illustrate our \textit{Look-Think-Answer} paradigm. In Case A, while the baseline suffers from object blindness, PointCoT explicitly scans the palm area to identify a cylindrical point cluster, correctly deducing the handheld object. In Case B, where side-view occlusion leads the baseline to hallucinate a standard windmill, our model strategically utilizes the Zenith View to verify a continuous manifold. Finally, instead of relying on semantic priors like the baseline, PointCoT grounds its reasoning in the Bottom View to verify a rectangular profile via sharp corner detection. These cases confirm that explicit rationale generation acts as a robust self-verification mechanism, effectively mitigating geometric hallucinations through targeted multi-view and 3D structural grounding.

\subsection{Ablation Studies}
We conduct comprehensive ablation studies to validate the necessity of each critical component within our framework, specifically focusing on the synergistic input modalities and the explicit reasoning mechanism.

\noindent{\textbf{\textit{Impact of Input Modalities.}}}
We first evaluate the dual-stream architecture by isolating geometric and semantic inputs. As shown in Table~\ref{tab:ablation_all}(a), single-modality variants yield suboptimal results. The image-only model struggles with fine-grained structural features (\eg, curvature) due to depth ambiguity, achieving the lowest accuracy. Conversely, the point-only model captures spatial topology but lacks the dense semantic texture required for robust recognition. By synergizing semantic richness with geometric precision, our hybrid architecture delivers a substantial +13.7\% gain over the point-only baseline. This confirms that integrating both modalities is indispensable for holistic 3D understanding.

\noindent{\textbf{\textit{Effectiveness of Chain-of-Thought.}}}
Next, we verify the hypothesis that explicit reasoning is superior to black-box mapping. As reported in Table~\ref{tab:ablation_all}(b), the \textit{Direct Mapping} baseline, despite utilizing dual-stream features, suffers from a high Hallucination Rate (25.4\%). This indicates that without explicit logical constraints, the model frequently relies on spurious semantic priors rather than verifiable 3D structures. While \textit{Implicit Reasoning} (training with rationales but skipping explicit generation during inference) offers marginal gains, the underlying decision process remains opaque. Our \textit{Explicit CoT} strategy, which forces the model to articulate geometry-grounded rationales before concluding, not only achieves the highest accuracy but also drastically reduces the Hallucination Rate to 5.1\%. This confirms that the explicit reasoning process acts as a robust self-verification mechanism to mitigate geometric hallucinations.

\section{Conclusion}
In this paper, we proposed PointCoT, the first framework to integrate explicit CoT reasoning into the 3D point cloud understanding pipeline. By enforcing a \textit{Look, Think, then Answer} mechanism, we demonstrated that explicit rationale generation serves as a powerful self-verification tool, significantly reducing hallucinations and enhancing interpretability.
Furthermore, to democratize research in this direction, we contributed Point-Reason-Instruct, a large-scale benchmark annotated with hierarchical reasoning chains. Extensive experiments validate that our approach not only achieves state-of-the-art performance but also exhibits superior data efficiency and zero-shot generalization.
Currently, our framework focuses on object-level reasoning. Extending this explicit reasoning paradigm to complex, cluttered indoor scenes or dynamic embodied manipulation tasks remains a promising frontier. We hope this work serves as a foundational step toward building truly transparent and intelligent 3D agents.

% \section*{Acknowledgments}
% \label{sec:ack}
% We thank xxx for their contributions and support for the project.

\clearpage
\newpage
\bibliographystyle{plainnat}
\setcitestyle{numbers}
\bibliography{ref}

% \clearpage
% \newpage
% \beginappendix

% \section{xxx}

\end{document}